\documentclass{article}

\usepackage{arxiv}

\usepackage[utf8]{inputenc} 
\usepackage[T1]{fontenc}    
\usepackage{hyperref}       
\usepackage{url}            
\usepackage{booktabs}       
\usepackage{amsfonts}       
\usepackage{nicefrac}       
\usepackage{microtype}      
\usepackage{graphicx}
\usepackage[numbers]{natbib}
\usepackage{doi}
\usepackage{float}
\usepackage{placeins}
\usepackage{needspace}
\usepackage{amsmath}
\usepackage{listings}
\usepackage{xcolor}
\usepackage{subcaption}
\usepackage{tcolorbox}

\usepackage{fancyhdr}
\tcbuselibrary{listings,skins,breakable}

\newtcblisting{promptbox}[1][]{
  colback=gray!5!white,
  colframe=gray!75!black,
  listing only,
  listing options={basicstyle=\ttfamily\small,breaklines=true},
  title={#1},
  breakable,
  enhanced,
  sharp corners,
  boxrule=0.5pt
}

\title{Extract-0: A Specialized Language Model for Document Information Extraction}

\author{
  Henrique Godoy \\
  Inteli \\
  São Paulo, Brazil \\
  henrique.godoy@sou.inteli.edu.br
}

\date{}

\hypersetup{
pdftitle={Extract-0: A Specialized Language Model for Document Information Extraction},
pdfsubject={cs.CL, cs.AI},
pdfauthor={Henrique Godoy}
}

\begin{document}
\maketitle

\begin{abstract}
This paper presents Extract-0 \footnote{Data and code available at \url{https://huggingface.co/datasets/HenriqueGodoy/extract-0} and \url{https://github.com/herniqeu/extract0}}, a 7-billion parameter language model specifically optimized for document information extraction that achieves performance exceeding models with parameter counts several orders of magnitude larger. Through a novel combination of synthetic data generation, supervised fine-tuning with Low-Rank Adaptation (LoRA), and reinforcement learning via Group Relative Policy Optimization (GRPO), Extract-0 achieves a mean reward of 0.573 on a benchmark of 1,000 diverse document extraction tasks, outperforming GPT-4.1 (0.457), o3 (0.464), and GPT-4.1-2025 (0.459). The training methodology employs a memory-preserving synthetic data generation pipeline that produces 280,128 training examples from diverse document sources, followed by parameter-efficient fine-tuning that modifies only 0.53\% of model weights (40.4M out of 7.66B parameters). The reinforcement learning phase introduces a novel semantic similarity-based reward function that handles the inherent ambiguity in information extraction tasks. This research demonstrates that task-specific optimization can yield models that surpass general-purpose systems while requiring substantially fewer computational resource.
\end{abstract}

\section{Introduction}

Document information extraction remains a critical bottleneck in enterprise automation, requiring the transformation of unstructured text into structured data according to predefined schemas. Organizations across healthcare, finance, legal, and regulatory sectors process millions of documents daily, yet current solutions require either extensive manual processing or deployment of large, computationally expensive language models that may not be feasible for many use cases.

Recent advances in large language models have demonstrated impressive capabilities in text understanding and generation. However, these general-purpose models often require hundreds of billions of parameters to achieve acceptable performance on extraction tasks, making them impractical for organizations with limited computational resources. Furthermore, their generalist nature means they may not achieve optimal performance on the specific task of structured information extraction.

This paper demonstrates that a targeted approach using a 7-billion parameter model can outperform models with one to two orders of magnitude more parameters on document extraction tasks. We achieve this through three key technical contributions: (1) a memory-preserving synthetic data generation pipeline that produces high-quality training examples while maintaining consistency across document chunks, (2) parameter-efficient fine-tuning that adapts only 0.53\% of model weights while achieving significant performance gains, and (3) a semantic similarity-based reward function for reinforcement learning that handles the inherent ambiguity in how information can be expressed in text.

Our approach treats information extraction as a well-defined transformation task with clear evaluation criteria. Given a document and an extraction schema, the model learns to identify and structure relevant information according to specified formats. This formulation enables systematic optimization through supervised learning followed by reinforcement learning, resulting in a specialized model that excels at its intended task.

Extract-0, our specialized extraction model, achieves a mean reward of 0.573 on a benchmark of 1,000 diverse document extraction tasks that were held out from the training data, significantly outperforming GPT-4.1 (0.457), o3 (0.464), and GPT-4.1-2025 (0.459). These benchmark tasks were selected from our synthetic data generation pipeline before model training commenced, ensuring unbiased evaluation. This performance advantage is achieved with a total training cost of only \$196, demonstrating that effective document extraction systems can be developed without massive computational budgets.

The key to Extract-0's performance lies in three technical innovations. First, we developed a synthetic data generation pipeline that produces 280,128 high-quality training examples from diverse document sources while maintaining consistency through a memory-preserving architecture. Second, we employ parameter-efficient fine-tuning using LoRA that modifies only 40.4M of the model's 7.66B parameters, enabling efficient adaptation without catastrophic forgetting. Third, we introduce a semantic similarity-based reward function that recognizes equivalent extractions despite surface form variations, addressing a fundamental challenge in training extraction models where the same information can be validly expressed in multiple ways.

The remainder of this paper is organized as follows. Section 2 presents the methodology, including the synthetic data generation pipeline, supervised fine-tuning configuration, and reinforcement learning approach. Section 3 analyzes experimental results, comparing Extract-0's performance against state-of-the-art models. Section 4 discusses implications for process automation and identifies limitations. Section 5 concludes with reflections on the broader impact of specialized extraction models. The appendix provides implementation details and hyperparameter configurations for reproducibility.

\section{Methodology}

\subsection{Data Collection and Document Sources}

This research employed a multi-source data collection strategy to ensure comprehensive coverage of diverse document types and extraction scenarios. The data collection infrastructure was designed to gather documents from four primary sources: arXiv for academic papers, PubMed Central for medical literature, Wikipedia for encyclopedic content, and FDA databases for regulatory documents. This diversification ensured that the resulting model would be capable of handling extraction tasks across various domains and document structures.

\subsection{Synthetic Data Generation Strategy}

The core innovation of this methodology lies in the synthetic data generation approach, which addresses the challenge of creating high-quality training data for document extraction tasks. The generation process employs a sequential memory-preserving architecture that maintains context across document chunks while generating extraction examples.

The mathematical formulation of the extraction task generation can be expressed as follows. For a document $D$ divided into chunks $\{c_1, c_2, ..., c_n\}$ where each chunk $c_i$ contains at most 2000 characters, the extraction function $E$ operates sequentially:

$$E(c_i) = f(c_i, M_{i-1})$$

where $M_{i-1}$ represents the accumulated memory of all extractions from previous chunks:

$$M_i = M_{i-1} \cup E(c_i)$$

This formulation ensures that information extracted from earlier chunks remains accessible when processing subsequent chunks, preventing contradictions and enabling the model to build upon previously identified entities and relationships. The memory preservation mechanism is crucial for maintaining consistency across long documents where related information may be distributed across multiple chunks.

The system processes documents using a hybrid parallel-sequential architecture. Within each document, chunks are processed sequentially to maintain contextual coherence, while multiple documents are processed in parallel to maximize throughput. This approach can be formally represented as:

$$\text{Processing} = \bigcup_{d \in D} \text{Sequential}(\{c_1^d, c_2^d, ..., c_n^d\})$$

where the union operation occurs in parallel across documents $d \in D$.

\begin{figure}[htbp]
\centering
\includegraphics[width=1.0\textwidth]{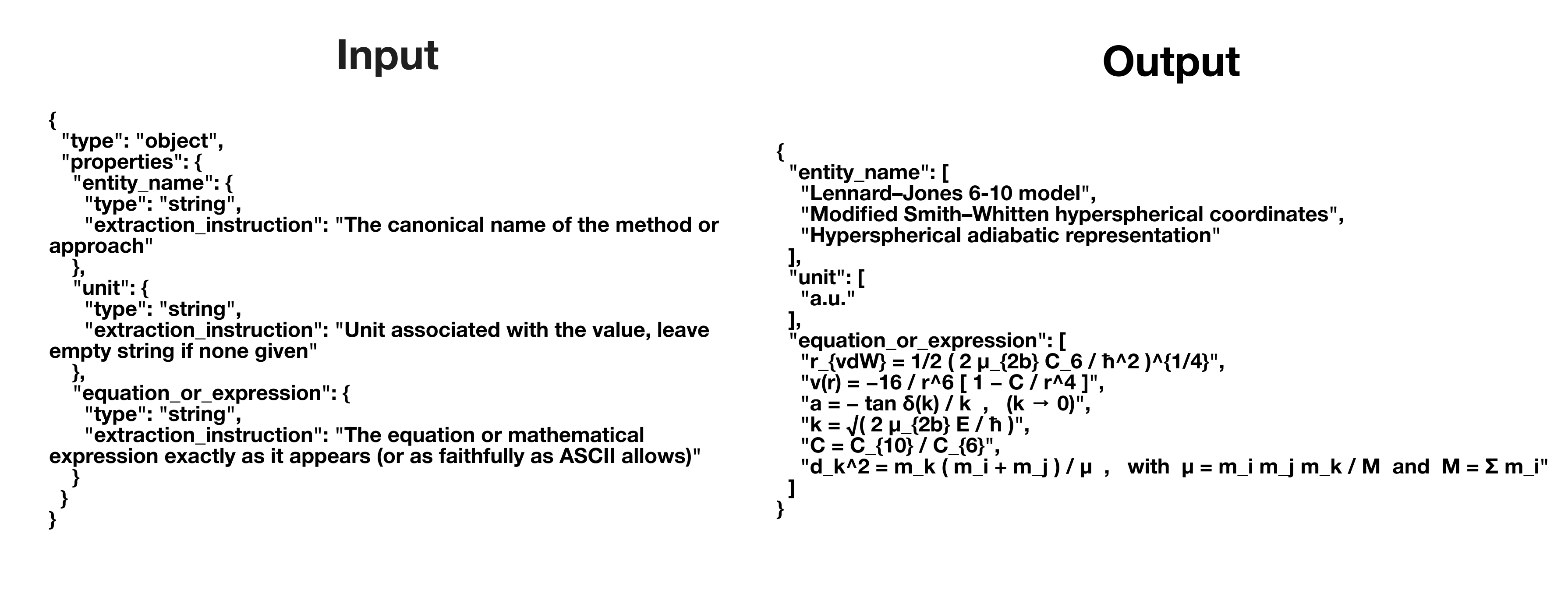}
\caption{Extraction task example: schema-guided transformation of unstructured text to structured JSON output}
\label{fig:extraction_example}
\end{figure}

Figure \ref{fig:extraction_example} illustrates the fundamental extraction task that the model learns to perform. Given an input schema defining the desired structure and a document containing unstructured text (the full document text is provided to the model although omitted from the figure for clarity), the model generates a JSON output that accurately captures the specified information. This example demonstrates extraction of scientific equations and entity names from technical documentation, showcasing the model's ability to preserve mathematical notation and handle nested object structures.

\subsection{Data Augmentation and Token Optimization}

The augmentation strategy was designed to create diverse training examples while maintaining computational efficiency. The system generates combinations of extracted fields from single documents and across multiple chunks, creating a rich variety of extraction scenarios. The augmentation process follows a controlled generation approach where the total token count for each training example (including prompt, schema, document chunks, and expected output) is constrained to remain within the range of 532 to 1900 tokens. During reinforcement learning, the maximum new tokens parameter is fixed at 532 (the lower bound) to ensure stable generation and prevent out-of-memory errors while maintaining sufficient output capacity for extraction tasks.

The token constraint was specifically chosen to avoid issues with sequence length limitations during supervised fine-tuning. The calculation of token count for each augmented example follows:

$$T(s, o, C) = |\text{Tokenizer}(\text{prompt} + \text{schema}_s + \bigcup_{c \in C} \text{chunk}_c + \text{output}_o)|$$

where $\text{Tokenizer}$ represents the tokenization function that converts text into tokens. During data preparation, token counting ensures that each training example fits within the model's context window. Note that in the reinforcement learning phase, generation is limited to a maximum of 532 new tokens, while the supervised fine-tuning data includes complete examples of 532-1900 total tokens. The function counts the total number of tokens after encoding the concatenated prompt, schema definition ($s$), document chunks ($C$), and expected output ($o$).

The augmentation process generates combinations through a probabilistic selection mechanism. For cross-chunk augmentation, which occurs with probability 0.7, the system selects 2-4 chunks with probabilities weighted as $\{0.5, 0.3, 0.2\}$ respectively. From each selected chunk, 1-3 fields are randomly sampled and combined to create a new training example. This approach resulted in the generation of 281,128 augmented examples, from which 1,000 were held out as a benchmark test set before training, leaving 280,128 examples for model training.

The schema combination process preserves the hierarchical structure of the extraction requirements. For a set of individual field schemas $\{S_1, S_2, ..., S_k\}$, the combined schema is constructed as:

$$S_{combined} = \{\text{type}: \text{object}, \text{properties}: \bigcup_{i=1}^{k} S_i\}$$

This ensures that the model learns to handle complex, multi-field extraction tasks while maintaining the ability to process individual fields accurately.

\subsection{Supervised Fine-Tuning Configuration}

The supervised fine-tuning phase employed Low-Rank Adaptation (LoRA) \cite{hu2021lora} to efficiently adapt the base DeepSeek-R1-Distill-Qwen-7B model \cite{deepseek2025r1} for extraction tasks. The training configuration utilized LoRA with rank $r=16$ and scaling factor $\alpha=32$, targeting both attention and MLP layers to maximize the model's capacity for learning extraction patterns. The adaptation can be mathematically expressed as:

$$W' = W_0 + \frac{\alpha}{r} BA$$

where $W_0$ represents the frozen pretrained weights, and $B \in \mathbb{R}^{d \times r}$ and $A \in \mathbb{R}^{r \times k}$ are the trainable low-rank matrices \cite{hu2021lora}. This configuration results in 40.4M trainable parameters out of the model's 7.66B total parameters, representing approximately 0.53\% of the model weights.

The training employed a constant learning rate schedule with warmup, where the learning rate $\eta(t)$ follows:

$$\eta(t) = \begin{cases} 
\frac{t}{T_{warmup}} \cdot \eta_{max} & \text{if } t < T_{warmup} \\
\eta_{max} & \text{if } t \geq T_{warmup}
\end{cases}$$

with $\eta_{max} = 10^{-4}$ and $T_{warmup}$ set to 8\% of total training steps. The training utilized mixed precision with bfloat16 to maintain numerical stability while reducing memory consumption, enabling effective batch sizes of 16 samples per gradient update.

The label masking strategy ensures that the model only receives gradient signals from the assistant's responses, not from the input prompts. For each training example, labels are constructed as:

$$\text{labels}_i = \begin{cases}
\text{token}_i & \text{if } i \geq \text{start}_{assistant} \\
-100 & \text{otherwise}
\end{cases}$$

where $\text{start}_{assistant}$ denotes the token position where the assistant's response begins. This selective masking prevents the model from memorizing input patterns and focuses learning on generation quality.

The training infrastructure employed gradient checkpointing to reduce memory consumption, allowing for larger effective context windows. The memory optimization strategy can be expressed as:

$$M_{required} = M_{model} + M_{gradients} \cdot \frac{1}{\sqrt{n_{checkpoints}}} + M_{activations} \cdot \frac{L}{n_{checkpoints}}$$

where $L$ represents the number of transformer layers \cite{vaswani2017attention} and $n_{checkpoints}$ the number of gradient checkpointing segments. This configuration enabled training with sequences up to 2048 tokens while maintaining stable memory usage under 4GB per device.

\subsection{Reinforcement Learning with Custom Reward Design}

The reinforcement learning phase presented a significant challenge in designing an appropriate reward function for document extraction tasks. Traditional metrics such as exact string matching prove inadequate for extraction tasks where multiple valid formulations may exist for the same information. To address this challenge, a novel reward function was developed based on field-level semantic similarity.

The reward function $R$ for an extraction output $y$ given a ground truth $y^*$ is formulated as a mean of field-wise similarities:

$$R(y, y^*) = \frac{1}{|\mathcal{F}|} \sum_{f \in \mathcal{F}} \text{FieldSim}(y_f, y^*_f)$$

where $\mathcal{F}$ represents the set of expected fields from the schema, and $\text{FieldSim}$ is a type-aware similarity function that handles different data types appropriately. The reward is set to zero if the extraction fails to produce valid JSON or lacks required fields, enforcing structural compliance as a hard constraint rather than a weighted component.

The field similarity function $\text{FieldSim}$ employs type-specific comparison strategies. For list-type fields containing multiple items, the similarity uses a bipartite matching approach:

$$\text{FieldSim}_{list}(P, G) = \frac{2 \cdot \sum_{(i,j) \in M} s_{ij}}{|P| + |G|}$$

where $P$ and $G$ are the predicted and gold lists respectively, and $M$ represents the optimal matching obtained by solving:

$$M = \arg\max_{M' \subseteq P \times G} \sum_{(i,j) \in M'} s_{ij} \quad \text{subject to} \quad s_{ij} > \tau$$

with threshold $\tau = 0.35$ to filter spurious matches. The pairwise similarity $s_{ij}$ between items is computed using cosine similarity of sentence embeddings \cite{reimers2019sentence} from a pre-trained MiniLM model \cite{wang2020minilm}:

$$s_{ij} = \frac{\text{embed}(p_i) \cdot \text{embed}(g_j)}{||\text{embed}(p_i)|| \cdot ||\text{embed}(g_j)||}$$

This formulation allows the reward function to recognize semantically equivalent extractions even when the exact wording differs, addressing the fundamental challenge that the same information can be expressed in multiple valid ways.

For scalar fields such as numbers and dates, the similarity function employs specialized handling. Numerical values use relative difference scoring, while dates are parsed and compared based on temporal distance. String fields that cannot be interpreted as dates utilize embedding-based semantic similarity through the sentence transformer model. This multi-modal approach ensures appropriate comparison regardless of data type.

\subsection{GRPO Training Architecture}

The reinforcement learning training employed a Group Relative Policy Optimization (GRPO) algorithm adapted for the extraction task domain. The policy update follows the clipped surrogate objective:

$$\mathcal{L}^{GRPO}(\theta) = \mathbb{E}_t \left[ \min\left( r_t(\theta) \hat{A}_t, \text{clip}(r_t(\theta), 1-\epsilon, 1+\epsilon) \hat{A}_t \right) \right]$$

where $r_t(\theta) = \frac{\pi_\theta(a_t|s_t)}{\pi_{\theta_{old}}(a_t|s_t)}$ is the probability ratio, $\hat{A}_t$ is the advantage estimate, and $\epsilon = 0.2$ is the clipping parameter.

The advantage estimation employs Generalized Advantage Estimation (GAE) \cite{schulman2016gae} with $\lambda_{GAE} = 0.95$:

$$\hat{A}_t = \sum_{l=0}^{\infty} (\gamma \lambda_{GAE})^l \delta_{t+l}^V$$

where $\delta_t^V = r_t + \gamma V(s_{t+1}) - V(s_t)$ represents the temporal difference error.

The training configuration for GRPO utilized a per-device batch size of 16 with 4 gradient accumulation steps, resulting in an effective batch size of 64. The learning rate was set to $5 \times 10^{-5}$ with no warmup, as the model was already pre-conditioned through supervised fine-tuning. The KL penalty coefficient was dynamically adjusted to maintain the KL divergence within the range $[1.5, 3.5]$ to balance exploration and stability:

$$\beta_{KL}^{(t+1)} = \begin{cases}
\beta_{KL}^{(t)} \cdot 1.5 & \text{if } D_{KL} < 1.5 \\
\beta_{KL}^{(t)} \cdot 0.5 & \text{if } D_{KL} > 3.5 \\
\beta_{KL}^{(t)} & \text{otherwise}
\end{cases}$$

This adaptive mechanism ensures that the model maintains sufficient exploration capability while preventing catastrophic divergence from the supervised fine-tuned baseline.

\section{Results}

\subsection{Model Performance Overview}

\begin{figure}[htbp]
\centering
\includegraphics[width=1.0\textwidth]{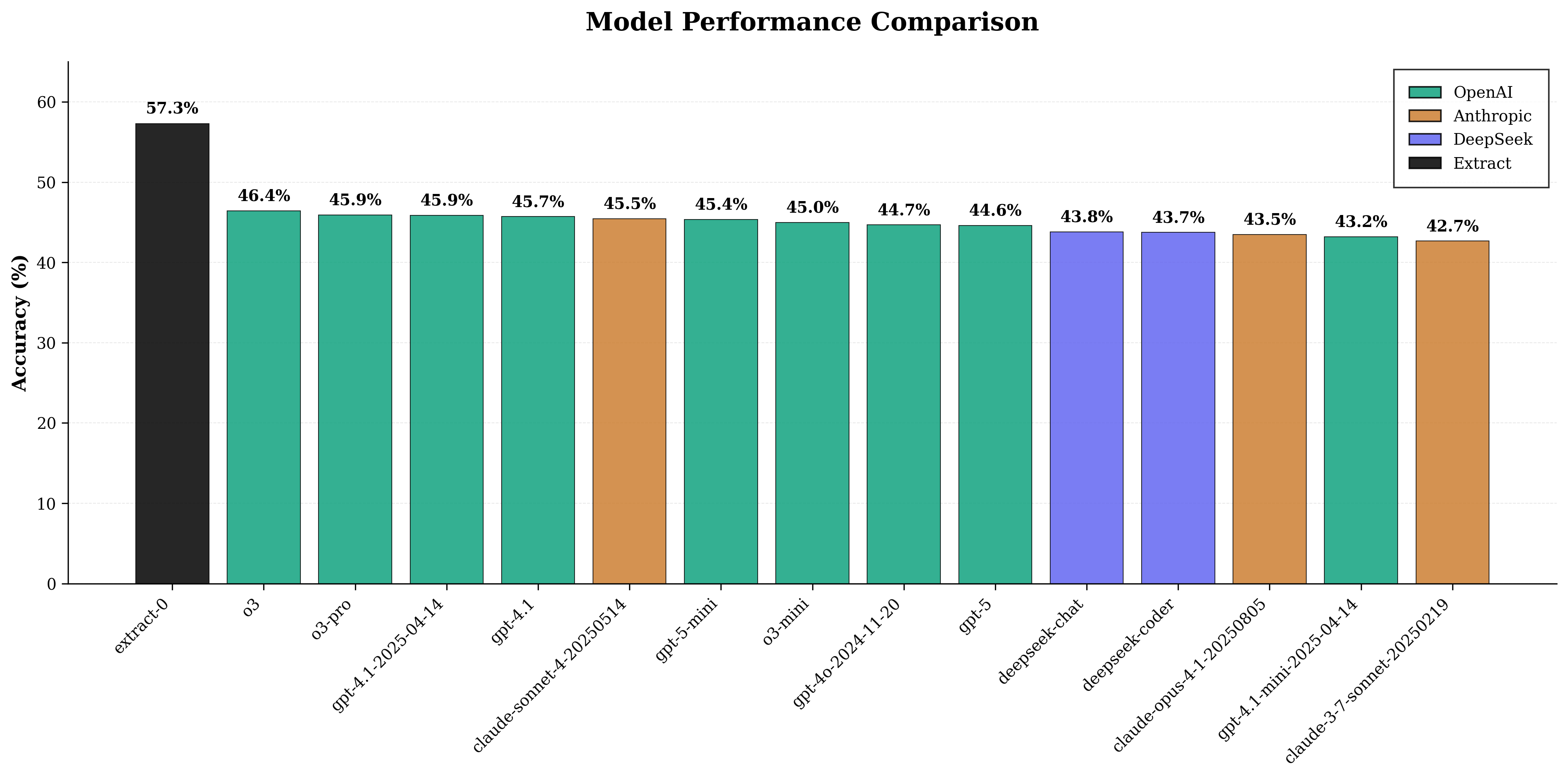}
\caption{Extract-0 achieves 0.573 mean reward, outperforming GPT-4.1 (0.457) and o3 (0.464) on 1,000 extraction tasks}
\label{fig:model_perfomance}
\end{figure}

Figure 1 (Model Performance Comparison) illustrates Extract-0's competitive positioning against established language models on our held-out benchmark. Despite being a specialized 7B parameter model, Extract-0 achieves a mean reward of 0.573 on the 1,000 test tasks that were excluded from both supervised fine-tuning and reinforcement learning phases, significantly outperforming larger general-purpose models including GPT-4.1 (0.457), o3 (0.464), and GPT-4.1-2025 (0.459). This performance advantage is particularly notable given that Extract-0 operates with substantially fewer parameters than these comparison models, demonstrating the effectiveness of task-specific optimization over general-purpose scaling.

\subsection{Training Dynamics Analysis}

\begin{figure}[htbp]
\centering
\includegraphics[width=0.7\textwidth]{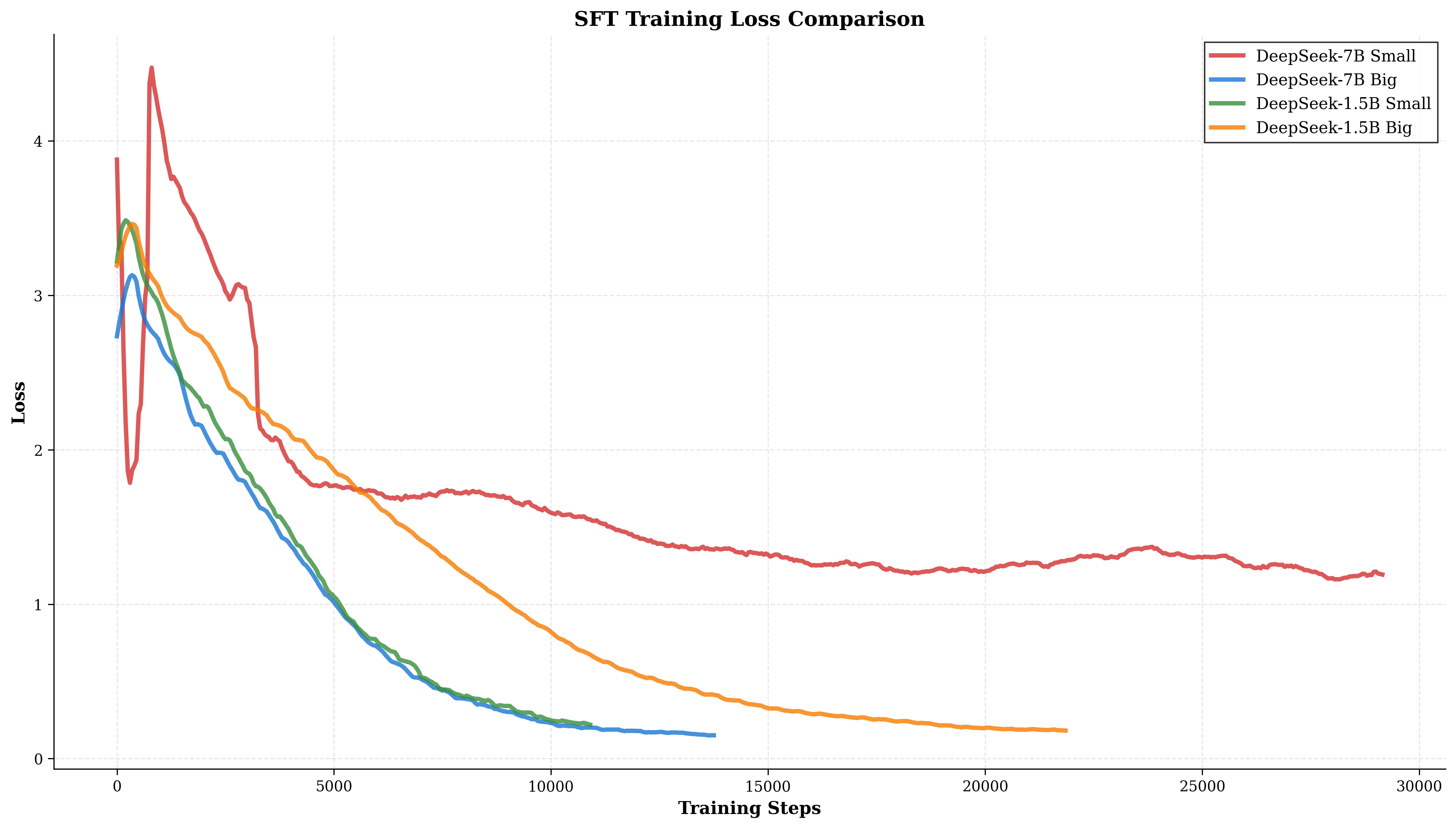}
\caption{SFT convergence: DeepSeek-7B variants reach asymptotic loss by 15k steps, large config achieves 0.2 final loss}
\label{fig:loss}
\end{figure}

Figure 2 (SFT Training Loss Comparison) presents the loss curves for model variants during supervised fine-tuning. The DeepSeek-R1-Distill-Qwen-7B configurations demonstrate rapid convergence within the first 5,000 training steps, with the large configuration achieving the lowest final loss of approximately 0.2. Smaller model variants show more gradual convergence patterns, with some configurations exhibiting initial instability before stabilizing. All models reach their asymptotic loss values by step 15,000, indicating efficient learning of the extraction task structure.

\subsection{Reinforcement Learning Optimization Results}

\begin{figure}[htbp]
\centering
\includegraphics[width=0.7\textwidth]{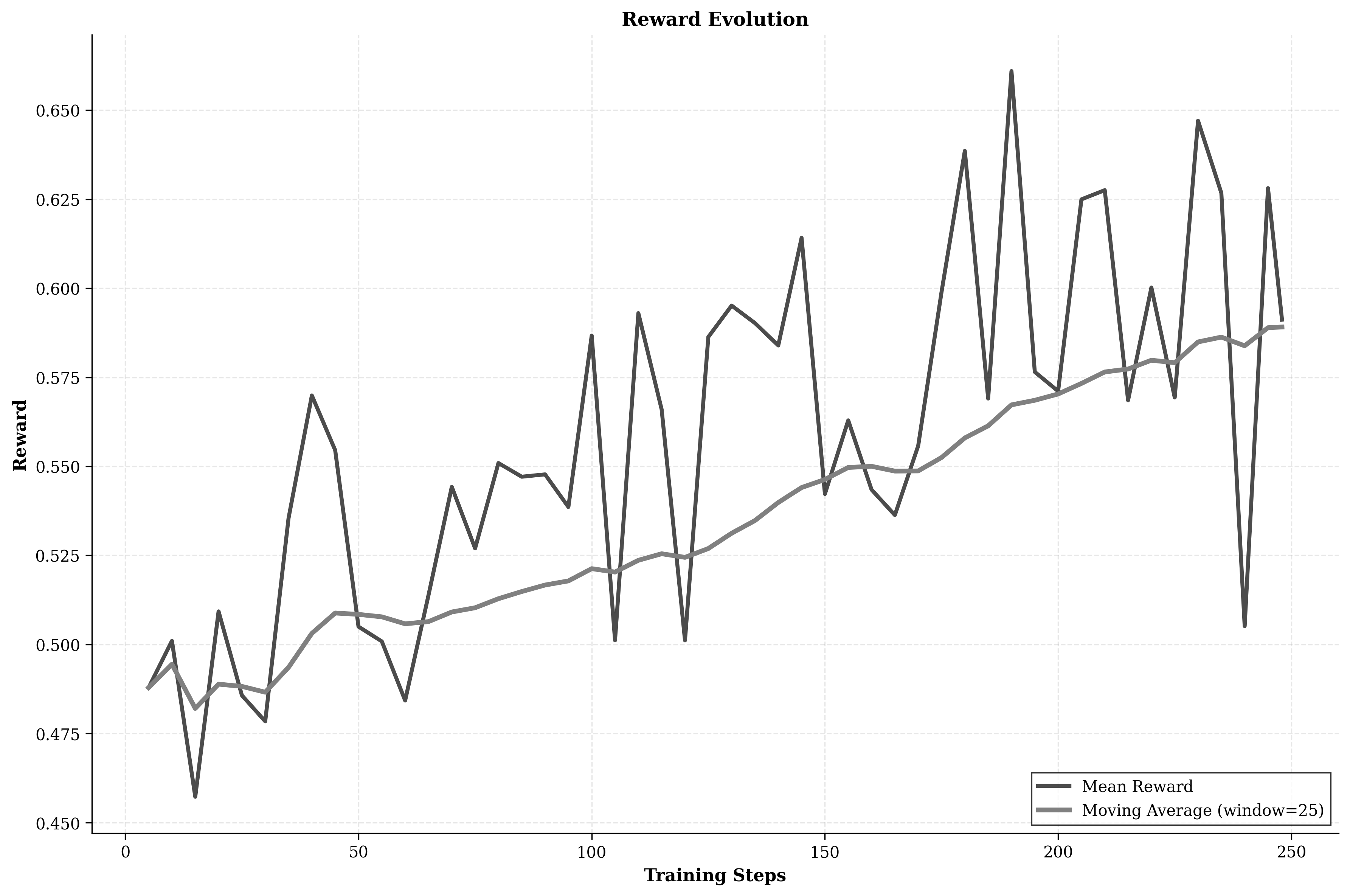}
\caption{GRPO training: mean reward improves 35.4\% from 0.488 to 0.661 peak over 248 steps}
\label{fig:reward}
\end{figure}

Figure 3 (GRPO Model 1: Reward Evolution) illustrates the reward progression during reinforcement learning optimization over 248 training steps. The mean reward trajectory shows consistent improvement from an initial value of 0.488 to a peak of 0.661 at step 190, representing a 35.4\% enhancement in extraction quality. The moving average trend line reveals three distinct phases in the optimization process.

The initial exploration phase (steps 0-50) exhibits high volatility with rewards fluctuating between 0.46 and 0.57, indicating active exploration of the policy space. The exploitation phase (steps 50-150) demonstrates steady improvement with the moving average increasing from 0.50 to 0.55, showing successful policy refinement. The convergence phase (steps 150-248) shows stabilization around 0.58-0.59 with occasional peaks reaching 0.66, suggesting the model has converged to a robust extraction policy while maintaining some exploration capacity.

\subsection{Comparative Performance Analysis}

Table 1 presents the comprehensive evaluation results across different training stages, demonstrating the incremental improvements achieved through each optimization phase. All evaluations were performed on the same 1,000 held-out tasks that were separated from the synthetic dataset before any model training began, ensuring no data leakage between training and evaluation.

\begin{table}[h]
\centering
\begin{tabular}{lccc}
\toprule
Model Configuration & Mean Reward & JSON Validity & Relative Improvement \\
\midrule
Base Model (DeepSeek-R1-Distill-Qwen-7B) & 0.232 & 42.7\% & Baseline \\
Base + SFT & 0.507 & 79.9\% & +118.5\% \\
Base + SFT + GRPO & 0.573 & 89.0\% & +147.0\% \\
\bottomrule
\end{tabular}
\caption{Performance comparison across training stages on 1,000 held-out extraction tasks}
\label{tab:performance}
\end{table}

The results demonstrate substantial performance gains at each optimization stage. The base DeepSeek-R1-Distill-Qwen-7B model without fine-tuning achieved a baseline mean reward of 0.232 with only 42.7\% of outputs producing valid JSON. After supervised fine-tuning, performance improved dramatically to 0.507 mean reward with 79.9\% JSON validity, representing a 118.5\% improvement over the baseline. The addition of reinforcement learning through GRPO further enhanced performance to 0.573 mean reward with 89.0\% JSON validity, yielding a cumulative improvement of 147.0\% over the base model.

\section{Discussion}

The results from Extract-0 provide evidence that task-specific optimization can compete with general-purpose scaling for specialized applications. The model's performance improvement of 25\% over GPT-4.1 (0.573 vs 0.457 mean reward) with substantially fewer parameters suggests that targeted approaches may offer advantages in resource-constrained environments where deploying larger models is economically or technically challenging.

The total training cost of \$196 (using a single H100 GPU) demonstrates that specialized extraction models can be developed with modest computational budgets. This cost structure makes custom model development accessible to a wider range of organizations, potentially enabling domain-specific extraction systems tailored to particular document types and requirements. The economic viability of this approach depends on the specific extraction needs and volume of documents processed by each organization.

Extract-0's architecture focuses on learning direct transformation functions from document text to structured data rather than general language understanding. In our benchmark of 1,000 extraction tasks, this approach achieved 89 \% JSON validity compared to the base model's 42.7\%, suggesting improved reliability for structured output generation. However, further evaluation on production workloads would be needed to confirm these benefits translate to real-world applications.

\subsection{Limitations and Future Directions}

Despite its strong performance, Extract-0 faces several limitations that warrant discussion. First, the model's training data, while diverse, may not capture all possible document formats and extraction scenarios. Highly specialized documents—such as technical patents, medical imaging reports, or financial derivatives contracts—may require additional fine-tuning for optimal performance. Additionally, the current implementation was trained exclusively on English-language documents, limiting its applicability to multilingual extraction tasks without further adaptation.

Second, the semantic similarity reward function, while more flexible than exact matching, may still miss nuanced extraction errors. For instance, extracting "John Smith" instead of "John P. Smith" receives a high similarity score despite the potential legal significance of the middle initial. Future work should explore hierarchical reward functions that weight extraction errors based on their downstream impact. An alternative approach could involve annotating a substantial corpus of extraction quality assessments and training a specialized language model to serve as a learned reward function, potentially capturing subtleties that rule-based similarity metrics overlook.

Third, the current implementation processes documents independently, without leveraging cross-document relationships. Many real-world extraction tasks involve entity resolution across multiple documents, requiring models to maintain consistent entity representations across a document corpus. Extending Extract-0 to handle multi-document extraction scenarios represents a natural evolution of the current approach.

\subsection{Broader Implications for AI Development}

The performance of Extract-0 on extraction tasks provides one data point in the ongoing discussion about specialized versus general-purpose models. Our results indicate that for the specific task of document information extraction, a 7B parameter specialized model can achieve competitive performance. Whether this pattern extends to other specialized tasks remains an open empirical question. The modular composition of such specialized models into larger systems presents both opportunities and challenges that warrant further investigation.

This modular approach offers several advantages over monolithic systems. First, it enables precise attribution of errors to specific components, facilitating targeted improvements. Second, it allows for graceful degradation—if one module fails, the system can route around it or invoke fallback mechanisms. Third, it supports regulatory compliance by making system behavior auditable at the component level.

\FloatBarrier
\section{Conclusion}

This paper presented Extract-0, a specialized 7B parameter model that achieves a mean reward of 0.573 on a benchmark of 1,000 held-out document extraction tasks, outperforming GPT-4.1 (0.457), o3 (0.464), and GPT-4.1-2025 (0.459). Through a combination of memory-preserving synthetic data generation (280,128 training examples plus 1,000 held-out test tasks), parameter-efficient fine-tuning (0.53\% of parameters modified), and semantic similarity-based reinforcement learning, Extract-0 demonstrates competitive performance with a training cost of \$196.

The results suggest that for specific tasks like document information extraction, specialized models can achieve strong performance without requiring the computational resources of larger general-purpose systems. Future work should evaluate whether these findings generalize to other specialized domains and assess performance on diverse production workloads. Additionally, investigating the composability of multiple specialized models and their integration into broader systems remains an important research direction.

\bibliographystyle{unsrtnat}

\begin{thebibliography}{6}

\bibitem[DeepSeek-AI(2025)]{deepseek2025r1}
DeepSeek-AI.
\newblock DeepSeek-R1: Incentivizing Reasoning Capability in LLMs via Reinforcement Learning.
\newblock \emph{arXiv preprint arXiv:2501.12948}, 2025.

\bibitem[Hu et al.(2021)]{hu2021lora}
Edward J. Hu, Yelong Shen, Phillip Wallis, Zeyuan Allen-Zhu, Yuanzhi Li, Lu Wang, and Weizhu Chen.
\newblock LoRA: Low-Rank Adaptation of Large Language Models.
\newblock \emph{arXiv preprint arXiv:2106.09685}, 2021.

\bibitem[Vaswani et al.(2017)]{vaswani2017attention}
Ashish Vaswani, Noam Shazeer, Niki Parmar, Jakob Uszkoreit, Llion Jones, Aidan N. Gomez, Łukasz Kaiser, and Illia Polosukhin.
\newblock Attention is All You Need.
\newblock In \emph{Advances in Neural Information Processing Systems}, 2017.

\bibitem[Reimers and Gurevych(2019)]{reimers2019sentence}
Nils Reimers and Iryna Gurevych.
\newblock Sentence-BERT: Sentence Embeddings using Siamese BERT-Networks.
\newblock In \emph{Proceedings of the 2019 Conference on Empirical Methods in Natural Language Processing (EMNLP)}, 2019.

\bibitem[Wang et al.(2020)]{wang2020minilm}
Wenhui Wang, Furu Wei, Li Dong, Hangbo Bao, Nan Yang, and Ming Zhou.
\newblock MiniLM: Deep Self-Attention Distillation for Task-Agnostic Compression of Pre-Trained Transformers.
\newblock In \emph{Advances in Neural Information Processing Systems}, 2020.

\bibitem[Schulman et al.(2016)]{schulman2016gae}
John Schulman, Philipp Moritz, Sergey Levine, Michael I. Jordan, and Pieter Abbeel.
\newblock High-Dimensional Continuous Control Using Generalized Advantage Estimation.
\newblock \emph{arXiv preprint arXiv:1506.02438}, 2016.

\end{thebibliography}

\appendix

\section{Implementation Details}

\subsection{Hardware Configuration}

All experiments were conducted on a single NVIDIA H100 80GB GPU. The total computational budget of \$196 was allocated as follows:
\begin{itemize}
\item Synthetic data generation: \$42
\item Supervised fine-tuning experiments: \$98
\item Reinforcement learning optimization: \$56
\end{itemize}

\subsection{Hyperparameter Configuration}

\subsubsection{Supervised Fine-Tuning}

\begin{verbatim}
Learning Rate: 1e-4
Batch Size: 16
Gradient Accumulation Steps: 1
Epochs: 5
Warmup Ratio: 0.08
Weight Decay: 0.01
LoRA Rank: 16
LoRA Alpha: 32
LoRA Dropout: 0.05
Target Modules: ["q_proj", "k_proj", "v_proj", "o_proj", 
                 "up_proj", "down_proj", "gate_proj"]
Mixed Precision: bfloat16
Gradient Checkpointing: Enabled
Max Sequence Length: 2048
\end{verbatim}

\subsubsection{Reinforcement Learning (GRPO)}

\begin{verbatim}
Learning Rate: 5e-5
Per-Device Batch Size: 16
Gradient Accumulation Steps: 4
Max Steps: 248
Temperature: 0.7
Top-p: 0.95
Beta (KL penalty): 0.05
Number of Generations per Prompt: 8
Max New Tokens: 532
Reward Scaling: "batch"
GAE Lambda: 0.95
Clip Range: 0.2
\end{verbatim}

\subsection{Data Processing Pipeline}

The synthetic data generation pipeline processes documents through the following stages:

\begin{enumerate}
\item \textbf{Document Chunking}: Split documents into 2000-character segments with 200-character overlap
\item \textbf{Sequential Extraction}: Process chunks sequentially, maintaining memory across chunks
\item \textbf{Augmentation}: Generate multi-field combinations with token count constraints (532-1900 tokens)
\item \textbf{Validation}: Ensure JSON validity and schema compliance for all training examples
\item \textbf{Stratification}: Balance training data across document types and extraction complexity levels
\end{enumerate}

\subsection{Evaluation Metrics}

The reward function implementation uses the following similarity thresholds:
\begin{itemize}
\item String similarity (embedding-based): Cosine similarity with MiniLM-L6-v2
\item Numerical similarity: Relative difference with 1.0 penalty for >100\% deviation
\item Date similarity: Temporal distance with exponential decay (half-life = 365 days)
\item List similarity: Bipartite matching with threshold $\tau = 0.35$
\item Boolean similarity: Exact match (1.0 for match, 0.0 for mismatch)
\end{itemize}

\subsection{Training Data Examples}

The following examples illustrate the diversity of extraction tasks in our synthetic training dataset:

\subsubsection{Example 1: Scientific Equation Extraction}

\begin{promptbox}[Schema]
{
  "type": "object",
  "properties": {
    "entity_name": {
      "type": "string",
      "extraction_instruction": "The canonical name of the method or 
        approach"
    },
    "unit": {
      "type": "string",
      "extraction_instruction": "Unit associated with the value, leave 
        empty string if none given"
    },
    "equation_or_expression": {
      "type": "string",
      "extraction_instruction": "The equation or mathematical 
        expression exactly as it appears (or as faithfully as ASCII allows)"
    }
  }
}
\end{promptbox}

\begin{promptbox}[Expected Output]
{
  "entity_name": [
    "Lennard-Jones 6-10 model",
    "Modified Smith-Whitten hyperspherical coordinates",
    "Hyperspherical adiabatic representation"
  ],
  "unit": [
    "a.u."
  ],
  "equation_or_expression": [
    "r_{vdW} = 1/2 ( 2 \\mu_{2b} C_6 / \\hbar^2 )^{1/4}",
    "v(r) = -16 / r^6 [ 1 - C / r^4 ]",
    "a = - tan \\delta(k) / k , (k \\to 0)",
    "k = \\sqrt( 2 \\mu_{2b} E / \\hbar )",
    "C = C_{10} / C_{6}",
    "d_k^2 = m_k ( m_i + m_j ) / \\mu , with \\mu = m_i m_j m_k / M 
      and M = \\Sigma m_i"
  ]
}
\end{promptbox}

\subsubsection{Example 2: Physics Relationship Extraction}

\begin{promptbox}[Schema]
{
  "type": "object",
  "properties": {
    "entity_name": {
      "type": "string",
      "extraction_instruction": "Give a concise descriptive name for the 
        physical relationship"
    },
    "equation_or_expression": {
      "type": "string",
      "extraction_instruction": "Write the equation exactly as it appears"
    },
    "context": {
      "type": "string",
      "extraction_instruction": "Provide the immediate textual context 
        supporting the claim"
    }
  }
}
\end{promptbox}

\begin{promptbox}[Expected Output]
{
  "entity_name": [
    "Collision-induced dimer dissociation rate",
    "Atom-dimer scattering cross section",
    "Atom-dimer relaxation rate",
    "van der Waals potential tail"
  ],
  "equation_or_expression": [
    "D_{ij}^{J\\Pi}=\\Sigma_{\\beta}(2J+1)\\pi/(\\mu_{k(ij)} k_{k(ij)}) |S_{\\beta\\leftarrow\\alpha_{ij}}^{J\\Pi}|^{2}",
    "\\sigma_{\\beta\\leftarrow\\alpha}^{J\\Pi}=(2J+1)\\pi/k_{\\alpha}^{2} |S_{\\beta\\leftarrow\\alpha}^{J\\Pi}-\\delta_{\\beta\\alpha}|^{2}",
    "\\beta_{k(ij)}^{J\\Pi}=\\Sigma_{\\beta}(2J+1)\\pi/(\\mu_{k(ij)} k_{k(ij)}) |S_{\\beta\\leftarrow\\alpha}^{J\\Pi}|^{2}",
    "v(r) \\to -C_6/r^6"
  ],
  "context": [
    "in a red-detuned optical dipole trap, the trapping potential is 
      proportional to the polarizability",
    "if the binding energy satisfies \\varepsilon_b < 6 U_trap, the dimer will 
      remain trapped",
    "dimers associated with a > 1300a0 have been shown to remain 
      trapped to some extent",
    "Elastic collisions with atoms can decelerate the dimer and lead 
      to the ejection of additional atoms",
    "These scattering processes are formally described by the 
      scattering matrix S"
  ]
}
\end{promptbox}

\subsubsection{Example 3: Financial Regulatory Extraction}

\begin{promptbox}[Schema]
{
  "type": "object",
  "properties": {
    "regulators": {
      "type": "array",
      "extraction_instruction": "List the regulators mentioned that 
        impose this requirement",
      "items": {
        "type": "string",
        "extraction_instruction": "Extract the name of each regulator"
      }
    },
    "name": {
      "type": "string",
      "extraction_instruction": "Full name of the regulator"
    },
    "event_description": {
      "type": "string",
      "extraction_instruction": "Extract the wording that describes 
        the risk event"
    },
    "description": {
      "type": "string",
      "extraction_instruction": "Extract the text that explains what 
        has changed or what the technology does"
    },
    "investor_type": {
      "type": "string",
      "extraction_instruction": "Extract the category label for the 
        investor type"
    }
  }
}
\end{promptbox}

\begin{promptbox}[Expected Output]
{
  "regulators": [
    ["Financial Conduct Authority (FCA)", 
     "U.S. Securities and Exchange Commission (SEC)"]
  ],
  "name": [
    "Financial Conduct Authority",
    "U.S. Securities and Exchange Commission"
  ],
  "event_description": [
    "Bank run with Bear Stearns"
  ],
  "description": [
    "Investment bank has considerable amounts of in-house software 
      created by the technology team, who are also responsible for 
      technical support",
    "Technology has changed considerably as more sales and trading 
      desks are using electronic processing",
    "Some trades are initiated by complex algorithms for hedging 
      purposes"
  ],
  "investor_type": [
    "Institutional investors",
    "Private investors"
  ]
}
\end{promptbox}

These examples demonstrate the variety of extraction patterns the model learns, from mathematical expressions and scientific relationships to bibliographic information. The training data includes 280,128 such examples spanning multiple domains and complexity levels.

\end{document}